# Hybrid ACO-CI Algorithm for Beam Design problems

**Ishaan R Kale[*1], Mandar S Sapre[2], Ayush Khedkar[2], Kaustubh Dhamankar[2], Abhinav Anand[2],**

**Aayushi Singh[2]**

[1]Institute of Artificial Intelligence, Dr Vishwanath Karad MIT World Peace University, Pune 411038, India

ishaan.kale@mitwpu.edu.in; kale.ishaan@gmail.com

[2]Symbiosis Institute of Technology, Symbiosis International (Deemed University), Pune 412115, India

mandar.sapre@sitpune.edu.in; ayush.khedkar.btech2019@sitpune.edu.in;

kaustubh.dhamankar.btech2019@sitpune.edu.in; abhinav.anand.btech2019@sitpune.edu.in;

aayushi.singh.btech2019@sitpune.edu.in

**Abstract**

A range of complicated real-world problems have inspired the development of several optimization methods. Here, a novel hybrid version of the Ant colony optimization (ACO) method is developed using the sample space reduction technique of the Cohort Intelligence (CI) Algorithm. The algorithm is developed, and accuracy is tested by solving 35 standard benchmark test functions. Furthermore, the constrained version of the algorithm is used to solve two mechanical design problems involving stepped cantilever beams and I-section beams. The effectiveness of the proposed technique of solution is evaluated relative to contemporary algorithmic approaches that are already in use. The results show that our proposed hybrid ACO-CI algorithm will take lesser number of iterations to produce the desired output which means lesser computational time. For the minimization of weight of stepped cantilever beam and deflection in I-section beam a proposed hybrid ACO-CI algorithm yielded best results when compared to other existing algorithms. The proposed work could be investigate for variegated real world applications encompassing domains of engineering, combinatorial and health care problems.

1. Introduction

It is well recognized that most of the real-world problems may not be solved analytically due to various drawbacks of traditional deterministic optimization methods like high computational cost, poor quality solutions and complex mathematical calculations. Additionally, there are several design constraints, objective functions, and different types of variables. Keeping all these factors in consideration the classical optimization algorithms are generally not an appropriate choice to solve such problems in spite of the fact that they provide exact solutions. Therefore, nature inspired optimization techniques have been used for handling a variety of optimization challenges like engineering and scientific trials including commercial decision-making, health-care and data analytics (Yang, 2020). The convergence rate, the processing time, the impartial exploitation and exploration, and the number of algorithm-specific control parameters get their design cues from nature. Several nature inspired metaheuristic algorithms such as Genetic Algorithm (GA) (Goldberg and Holland, 1988), Particle Swarm Optimization (PSO) (Eberhart and Kennedy, 1995), Ant Colony Optimization (ACO) (Dorigo and Gambardella, 1997), Firefly Algorithm (FA) (Yang 2009), Cuckoo Search Algorithm (CS) (Feng et al.,2014), Artificial Bee Colony Optimization (ABC) (Karaboga, 2005), etc. are quite effective to solve complex real-world applications. These algorithms have shown adequate problem-solving ability. However, these algorithms might perform very well for some problems while it may perform poorly for others. This may be due to the characteristics of these algorithm being suitable for the particular set of problems. The metaheuristic algorithms are not able to explore the search space in the case of discrete and mixed design variables problems. In contrast, nature inspired algorithms provide the best solution for a variety of problems in variegated areas in a respectably shorter amount of computing time as compared to traditional optimization techniques.

Thus, every algorithm has some advantages and some limitations. In order to overcome it the key features of two or more algorithms can be merged to get the better version of the algorithm. The hybrid algorithm may be developed by using features of one algorithm to overcome the limitations of the second algorithm and vice-versa. Introduced by Dorigo and Gambardella (1997), ACO is a metaheuristic algorithm which is inspired by the foraging behaviour of ants. Whereas, Cohort Intelligence (CI) is a socio-inspired metaheuristic that was put forth by (Kulkarni et al., 2017; Kulkarni et al., 2013) and is based on the candidates in a cohort's self-supervised learning behaviour. The present work is an attempt to investigate hybridization of ACO and Cohort Intelligence (CI).

## 2. Survey of the Parent Algorithms

### 2.1 Ant Colony Optimization

Dorigo and Gambardella (1997) first developed the ACO as a type of simulative evolutionary algorithm, which was influenced by the foraging behaviour of ants in nature. When an ant is out foraging and encounters an obstacle on the road they have never been on before, they will randomly choose one path and secrete pheromones to help other ants decide which way to take. A path's likelihood of being used by other ants increases as more pheromones are deposited along it. Because of this, the pheromone trail along such a path will build up quickly and draw in additional ants (a process known as positive feedback) (Tsai et al., 2010). Based on this natural process, ant colonies arrive at the best answer by sharing information and working together, all without any prior knowledge. The advantages of parallel computation, self-learning, and efficient information feedback makes ACO as an effective intelligence-based problem-solving methodology. In the beginning of the search process the information is scarce which affects convergence rate. ACO algorithm was used to solve various NP-hard combinatorial optimization problems like vehicle routing, travelling salesman problems and dynamic continuous problems (Stützle and Dorigo, 1999). Further, it is applied to solve the problems from structural engineering and design engineering domain (Mohan and Baskaran, 2012).

The performance of ACO algorithm is improved by hybridizing it with other contemporary algorithms. For e.g., ACO-PSO which was introduced (Luan *et al.*, 2019) where the search space is expanded by local exploration and the search process is directed by the global experience. The PSO method is used to determine the optimum values for the parameters, which are required in the ACO algorithm's city selection procedures and define the importance of inter-city pheromone and distances for Traveling Salesman problem. The 3-Opt algorithm is used to enhance city selection processes that the ACO algorithm was unable to enhance due to local minimums dropping below thresholds (Mahi et al., 2015).

A hybrid algorithm of ant colony optimization and artificial bee colony optimization (ACO-ABC) (Kefayat et al., 2015). is proposed to solve placement and sizing of distributed energy resources (DERs) in an optimized way. It uses ABC's discrete structure technique to optimize location and ACO's continuous structure technique to optimize size. This is done to achieve advantages of the global and local search ability of both the individual algorithms. The hybrid ACO-CS (Jona and Nagaveni, 2014) is based on swarm to perform feature selection in Digital Mammogram. ACO algorithm is also hybridized with taboo search algorithm (Huang and Liao, 2008) to solve classical job shop scheduling problems. The algorithm incorporates a novel decomposition method inspired by the shifting bottleneck procedure, as well as a mechanism of occasional re-optimizations of partial schedules, in place of the traditional construction strategy to produce workable schedules. Additionally, a taboo search method is integrated to enhance the quality of the solutions.

A hybrid ACO (HACO) for the Next Release Problem (NRP) (Jiang et al., 2010) is a NP-hard problem where the goal is to balance the customer demands, resource limitations, and requirement dependencies. Multiple artificial ants are used, to build new solutions. Additionally, a local search is added to HACO to enhance the quality of the solutions (first discovered when hill climbing). The experimental findings showed that HACO have shown better performance than ACO algorithms in terms of computational time and solution quality. A hybridization of ACO with Simulated Annealing referred to as ACO-SA is proposed by Dengiz et al. (2010) for designing the communication networks. Finding the best network architecture with the lowest overall cost and the highest degree of dependability across all terminals is the design challenge. The hybrid ACO-SA utilizes the ability of ACO to locate higher performance solutions and the capacity of SA to leave local minima and find superior solutions. ACO has also been hybridized with Genetic Algorithm to solve protein function prediction and text feature selection (Basiri and Nemati, 2009; Nemati et al., 2009). The GA-ACO-PSO hybrid algorithm (Tam et al., 2018) is introduced to address various issues in optimization process. Its viability has been tested using a variety of unconstrained multimodal and unimodal test functions, and the suggested hybrid algorithm outperforms more established GA, ACO, and PSO in terms of repeatability and accuracy.

**2.2. Cohort Intelligence**

Cohort Intelligence (CI) is a socio-inspired metaheuristic conceptualized by Kulkarni et al. (2013) is based on the self-supervised learning approach of the candidates in a society. Every candidate repeatedly tries to emulate peers' behaviour in order to improve its own behaviour. Kulkarni and Shabir (2016) employed CI to resolve combinatorial challenges, including the well-known Traveling Salesman Problem (TSP) and the 0-1 Knapsack Problem. In order to address an emerging healthcare issue, Kulkarni et al., (2016) utilized CI to develop a cyclic surgical schedule that minimized bottleneck in the recovery unit. Additionally, it was employed to address issues with cross-border transit. Sarmah and Kulkarni (2017, 2017a) spoke about two steganographic methods utilising CI with Modified Multi-Random Start (MMRS) and Cognitive Computing (CC) Local Search employing Joint Photographic Expert Group (MMRSLS) Greyscale picture with (JPEG) compression applied to cover up text. The cryptography algorithms based on CI was developed by Sarmah and Kale (2018). Additionally, binary optimization issues were demonstrated to be amenable to CI (Aladeemy et al., 2017). The CI algorithm is also investigated for solving various problems in mechanical engineering domain like truss structure, design engineering and manufacturing domain (Kale and Kulkarni, 2017; Kale et al., 2019; Kale and Kulkarni, 2021; Kale et al, 2022). The economic optimization of shell and tube heat exchanger was discussed by Dhavle et al. (2018). The CI algorithm is applied for mesh-smoothing of the hexahedral elements (Sapre et al., 2018). Several variations of this approach were proposed by (Patankar and Kulkarni, 2018) and assessed over seven multimodal and three unimodal unconstrained test functions. For the smoothing of hexahedral mesh in cubical and prismatic geometries, Sapre et al. (2019) employed variants of CI. The Multi CI method created by (Shastri and Kulkarni, 2018) focuses on similar and cross functional learning processes among many cohorts.

The algorithm's tendency is to follow each other exclusively during the exploration phase results in premature convergence. This is overcome in hybrid algorithm referred to as K-means with modified CI (K-MCI) (Krishnasamy et al., 2014). The CI algorithm is hybridized with Colliding Bodies Optimization (CBO) incorporated with Self-Adaptive Penalty Function (SAPF) approach referred to as CI-SAPF-CBO for solving the convex constrained optimization problems arising in truss structure domain, design engineering domain, manufacturing domain (Kale and Kulkarni, 2021), industrial and chemical process, process design and synthesis, power system, power electronics and livestock feed ration optimization (Kale and Kulkarni, 2023). The CI-SAPF-CBO was developed to eliminate the sampling space reduction factor. The Adaptive Range GA (Iyer et al., 2019) is a hybrid algorithm of GA while CI is used to make the mutation process self-adaptive. It is applied to the economic optimization of shell and tube heat exchanger design problem. CI algorithm is combined with the mean value theorem to develop the procedures for stiffness matrices using numerical integration (Sapre et al., 2023).

## 3. ACO-CI Hybrid Algorithm

The proposed algorithm is a combination of ACO and CI to generate a hybrid algorithm ACO-CI to obtain optimized solution for mechanical design problems when compared with results obtained by contemporary algorithms. In the proposed approach, the process starts by first setting the computational parameter of CI i.e., cohort size and reduction factor. The parameters of ACO i.e., number of ants, constant parameters, dimension, initial solution, evaporation rate, pheromone level. The likelihood of the path is then chosen in accordance with the pheromone level. Assumed that five best ants are considered, following that, the function values of each ant are assessed, and one best and four better ants are chosen based on the function values. Then, using a roulette wheel technique, the odds of these five ants are determined, and the best ant is followed by better ants. These five ants are combined with the rest of the ant population after the operation is finished. The best ant's sampling area is chosen, and the same is updated for the remaining ants. Convergence is examined when the cycle is complete, and if the convergence is not reached, the procedure is repeated to determine the likelihood of choosing the best option; otherwise, the present answer is accepted as the outcome.

### 3.1 Mathematical Modelling

The main structure of the proposed hybrid ACO-CI algorithm is presented below. The mathematical description of ACO-CI is explained considering a general unconstrained optimization problem (in minimization sense) as follows:

Minimize $F(X) = F(x_1, \ldots, x_i, \ldots, x_N)$ (1)

Subjected to $\Phi_i^{lower} < x_i < \Phi_i^{upper}, i = 1,2,3, \ldots n$ (2)

**STEP 1:**

Considering the number of ants as $a$ where each individual ant $a$ ($a = 1,2,3, \ldots, a$) containing a set of variables $w = (w_1, w_2, w_3, \ldots, w_m)$. The initial solution is randomly generated similar to the other population-based technique as follows:

$x = (\Phi_i^{upper} - \Phi_i^{lower}) \times rand(a, w) + \Phi_i^{lower}$ (3)

**STEP 2:**

Defining the probability of path selection and then calculating cumulative probability ranges associated with each path. The probability is calculated using the initial pheromone level which is given by

$P_A(x) = \frac{\tau_{(i)}}{a}$ (4)

where $\tau_{(i)}$ = initial pheromone level ($\tau_{(i)} = 1$)

Generating random values in range (0, 1) for each ant for $a$ ($w$). The corresponding search space values assigned to the cumulative probability range is substituted in the function (**X**) as mentioned in equation (1) to find the minimum and maximum values for the same

**STEP 3:**

The function values were then arranged in ascending order from which the five most minimum values were selected so that they can be taken into consideration for further selection

**STEP 4:**

The probability of selecting path $F(X)$ of every associated ant $a$ ($a = 1, 2, \ldots, a$) is calculated as follows:

$P_c(X) = \frac{\frac{1}{F(X)}}{\sum_{a=1}^{N} 1/F(X)}$ (5)

Using roulette wheel approach, each ant decides to follow the corresponding path and associated attributes.

**STEP 5:**

Every candidate $a(a = 1, 2, \ldots, a)$ shrinks the sampling interval $r_i$ ($i = 1,2,3, \ldots, n$) associated with every variable $W_i$ ($i = 1,2,3, \ldots, n$) to its local neighbourhood. This is done as follows

$range = (\Phi_i^{upper} - \Phi_i^{lower})$ (6)

$r_{new} = r \times \frac{range}{2}$ (7)

This new range is utilized to calculate the new lower and upper bound for the further iteration.

**STEP 6:**

For further iteration, the new updated pheromone $\tau$ level for best and other ants is calculated using the following formula

$$\tau_{(i)} = \tau_{(i-1)} + Q \times \left(\frac{F_{best(i-1)}}{F_{worst(i-1)}}\right) \tag{8}$$

$$\tau_{other} = (1-\rho) \times \tau_{best} \tag{9}$$

Where $Q$ = constant parameter

$\rho$ = evaporation rate

The new lower and upper bound are then used to formulate a new search space which is then used in the subsequent iteration till convergence is achieved

**STEP 7:**

Upon achieving convergence, the following conditions are evaluated:

$$F_{best} = F_{worst} \tag{10}$$

When the values of $F_{best}$ and $F_{worst}$ are same, then we can conclude that convergence is achieved

The flowchart of proposed hybrid ACO-CI is presented in Figure 1.

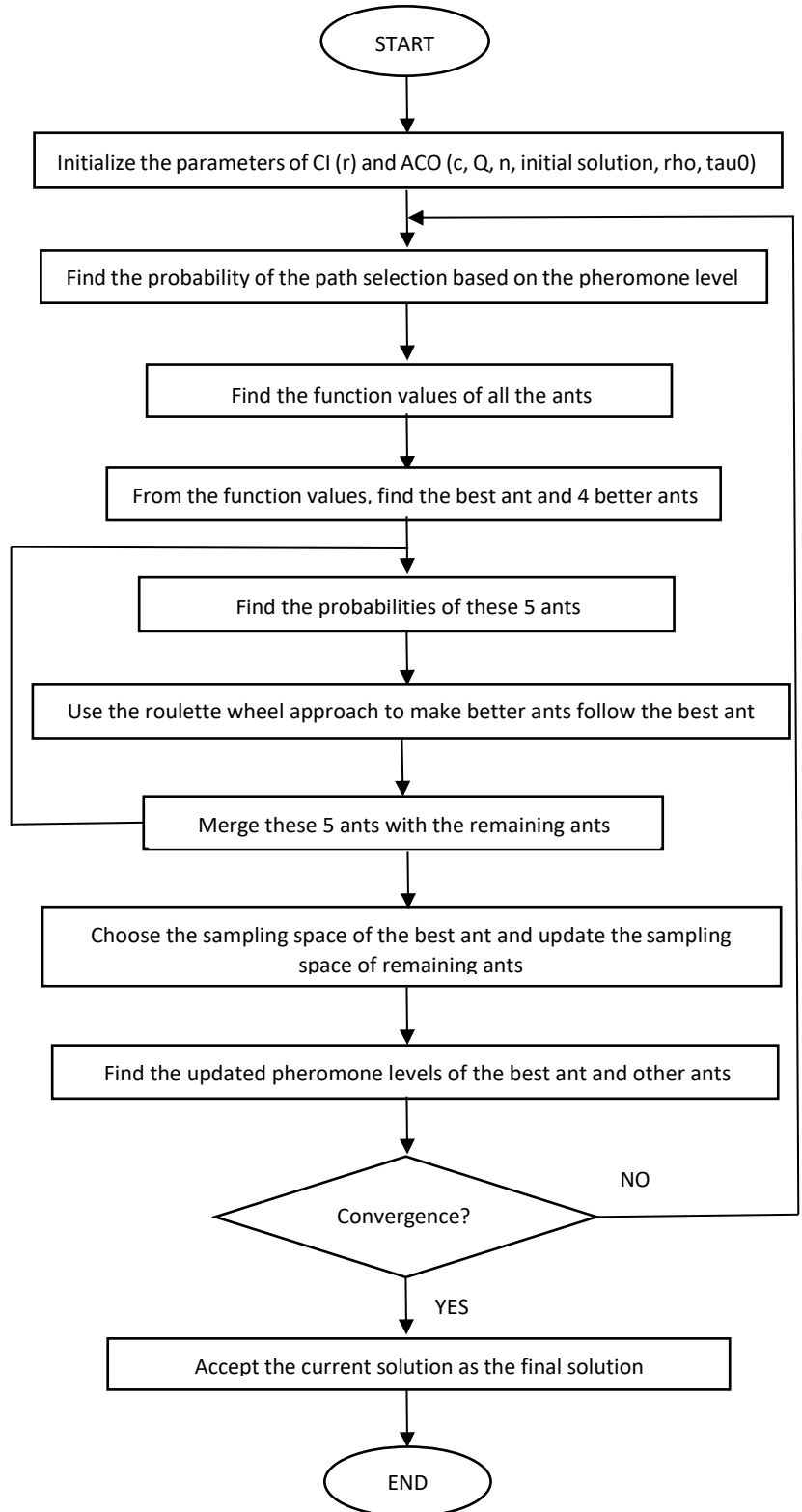

Fig 1. Hybrid ACO-CI Flowchart

## 4. Comparison and Analysis

### 4.1 Statistical analysis

**Table. 1** Statistical comparison of results obtained by ACO-CI with PSO, ABC, BSA, GA, CI, ARGA

| Sr. No | Results | PSO-2011 (Hariya, 2016) | ABC | BSA (Patterson, et al., 1990) | GA (Iyer et al., 2019) | CI (Iyer et al., 2019) | ARGA (Iyer et al., 2019) | ACO-CI |
|---|---|---|---|---|---|---|---|---|
| F2 | Mean | 3 | 3.000000047 | 3 | 3.002654786 | 3.060050207 | 3 | 12.0615 |
|  | Std | 0 | 0 | 0 | 0.0071 | 0.0696 | 0 | 8.781 |
|  | Best | 3 | 3 | 3 | 3.000001498 | 3.001805304 | 3 | 3.1298 |
| F5 | Mean | 1.521432297 | 3.4E-14 | 1.05E-14 | 0.000424243 | 2.4128E-05 | 0 | 4.44E-16 |
|  | Std | 0.6618 | 0 | 0 | 0.0006 | 0 | 0 | 3.01E-31 |
|  | Best | 8E-15 | 2.93E-14 | 8E-15 | 2.25229E-05 | 5.66423E-06 | 0 | 4.44E-16 |
| F6 | Mean | 4.1923E-09 | 2.8E-15 | 0 | 2.91934E-06 | 0.398107773 | 0 | 0.7747 |
|  | Std | 0 | 0 | 0 | 0 | 0.4036 | 0 | 0.6032 |
|  | Best | 0 | 5E-16 | 0 | 4.63631E-05 | 0.004657527 | 6.53209E-06 | 0.0010 |
| F7 | Mean | 0 | 0 | 0 | 6.63584E-05 | 9.88924E-05 | 0 | 2.12E-12 |
|  | Std | 0 | 0 | 0 | 0.0019 | 0.0035 | 0 | 2.40E-12 |
|  | Best | 0 | 0 | 0 | 0.006907811 | 0.003961277 | 1.18985E-10 | 9.99E-14 |
| F8 | Mean | 0 | 0 | 0 | 0.218317239 | 2.47393E-05 | 0 | 2.41E-12 |
|  | Std | 0 | 0 | 0 | 0.2185 | 0.0051 | 0 | 3.14E-12 |
|  | Best | 0 | 0 | 0 | 0.000545683 | 0.004768095 | 0 | 1.38E-14 |
| F9 | Mean | 0 | 6E-16 | 0 | 0.002793445 | 0.000116954 | 2.47393E-05 | 5.83E-13 |
|  | Std | 0 | 0 | 0 | 0.0253 | 0.0035 | 0.0051 | 5.28E-13 |
|  | Best | 0 | 1E-16 | 0 | 0.051830466 | 0.003944181 | 0.004768095 | 9.16E-15 |
| F10 | Mean | 0 | 0 | 0 | 2.02177E-09 | 0.000111086 | 0 | 74 |
|  | Std | 0 | 0 | 0 | 0 | 0.0053 | 0 | 0 |
|  | Best | 0 | 0 | 0 | 0.000104348 | 0.004105462 | 0 | 74 |
| F11 | Mean | 0.397887358 | 0.397887358 | 0.397887358 | 0.397887393 | 0.397887393 | 0.397887358 | 1.2429 |
|  | Std | 0 | 0 | 0 | 0.3979 | 0.3979 | 0.3979 | 0.9956 |
|  | Best | 0.397887358 | 0.397887358 | 0.397887358 | 1.37389E-07 | 1.37389E-07 | 1E-16 | 0.4033 |
| F13 | Mean | 0.666666667 | 3.8E-15 | 0.644444444 | 51485.41983 | 5.973259962 | 9.18261E-09 | 0.2639 |
|  | Std | 0 | 0 | 0.1217 | 91025.788 | 184.0608 | 0.0021 | 0.1665 |
|  | Best | 0.666666667 | 2.1E-15 | 0 | 26459.09573 | 255.9487027 | 0.00928228 | 0.0350 |
| F14 | Mean | −1.0000000000000000 | −1.0000000000000 | −1.0000000000000 | −0.0000010654169759 | −0.0000010654169759 | −1.0000000000000000 | -3.00E-09 |
|  | Std | 0 | 0 | 0 | 0 | 0 | −1.0000 | 1.30E-24 |
|  | Best | −1.0000000000000000 | −1.0000000000000 | −1.0000000000000 | 2.39064E-07 | 2.39064E-07 | 1.0856E-05 | -3.00E-09 |
| F18 | Mean | 0.006894369 | 0 | 0.000493069 | 3.70323E-06 | 0.000514025 | 1E-16 | 0 |
|  | Std | 0.0081 | 0 | 0.0019 | 0 | 0.0005 | 0 | 0 |
|  | Best | 0 | 0 | 0 | 8.01584E-08 | 1.91539E-05 | 0 | 0 |
| F19 | Mean | −3.8627821478207500 | −3.8627821478207500 | −3.8627821478207500 | −3.8627821341187400 | −3.8622838932951700 | −3.8627821478207600 | -3.3308 |
|  | Std | 0 | 0 | 0 | −3.8628 | −3.8569 | −3.8628 | 0.3719 |
|  | Best | −3.8627821478207600 | −3.8627821478207600 | −3.8627821478207600 | 9.90083E-07 | 0.005275084 | 1.56749E-09 | -3.7948 |
| F20 | Mean | −3.3180320675402500 | −3.3219951715842400 | −3.3219951715842400 | −3.3219951655723100 | −3.3059856369810300 | −3.3219951715697400 | -2.5068 |
|  | Std | 0.0217 | 0 | 0 | −3.3220 | −3.2587 | −3.3220 | 0.1639 |
|  | Best | −3.3219951715842400 | −3.3219951715842400 | −3.3219951715842400 | 2.64744E-06 | 0.037140454 | 3.00948E-08 | -2.882 |
| F21 | Mean | 0.000307486 | 0.000441487 | 0.000307486 | 0.005707216 | 0.000368907 | 0.000307494 | 0.0118 |
|  | Std | 0 | 0.0001 | 0 | 0.0218 | 0.0006 | 0.0003 | 0.0094 |
|  | Best | 0.000307486 | 0.000323096 | 0.000307486 | 0.010740875 | 0.000140003 | 6.95643E-05 | 0.0004 |
| F23 | Mean | −1.3891992200744600 | −1.4999990070800800 | −1.4821658762555300 | −1.4999999907728900 | −0.7976938208317790 | −1.5000000000000000 | -1.3195 |
|  | Std | 0.2257 | 0 | 0.0977 | −1.5000 | −0.1543 | −1.5000 | 0.2489 |
|  | Best | −1.4999992233524900 | −1.4999992233524900 | −1.4999992233524900 | 2.64567E-06 | 0.278211478 | 2.8958E-08 | -1.5666 |
| F24 | Mean | −0.9166206788680230 | −0.8406348096500680 | −1.3127183561646500 | −1.4991682175725200 | −0.0023646048023792 | −1.4999999488423700 | -1.5 |
|  | Std | 0.3918 | 0.2001 | 0.3159 | 0.0018 | 0.005 | 0 | 0 |
|  | Best | −1.5000000000003800 | −1.4999926800631400 | −1.5000000000003800 | −1.4999934260674600 | −0.0235465240654852 | −1.5000000000000000 | -1.5 |
| F25 | Mean | 0 | 4E-16 | 0 | 7.91055E-07 | 0.001670295 | 3.22909E-07 | 0.0120 |
|  | Std | 0 | 0 | 0 | 0 | 0.0026 | 0 | 0.0119 |
|  | Best | 0 | 1E-16 | 0 | 4.33045E-07 | 3.46951E-07 | 0 | 0.0002 |
| F26 | Mean | −1.8210436836776800 | −1.8210436836776800 | −1.8210436836776800 | −1.8036302197863400 | −1.8092292166278800 | −1.8210436447465000 | -1.6917 |
|  | Std | 0 | 0 | 0 | 0.0185 | 0.0149 | 0 | 0.0805 |
|  | Best | −1.8210436836776800 | −1.8210436836776800 | −1.8210436836776800 | −1.8205127535579800 | −1.8210355169086100 | −1.8210436835996600 | -1.8203 |
| F27 | Mean | −4.6565646397053900 | −4.6934684519571100 | −4.6934684519571100 | −4.5660594921319500 | −4.3603011700638300 | −4.6934684094269700 | -2.5580 |
|  | Std | 0.0557 | 0 | 0 | 0.0653 | 0.3016 | 0 | 0.3309 |
|  | Best | −4.6934684519571100 | −4.6934684519571100 | −4.6934684519571100 | −4.6871906135714200 | −4.6401791039267200 | −4.6934684515139900 | -3.2790 |
| F30 | Mean | 1.30719E-05 | 0.000260433 | 2.84432E-09 | 0.046618097 | 153.1867735 | 0.000193344 | 4.10E-15 |
|  | Std | 0 | 0 | 0 | 0.1788 | 104.147 | 0.0001 | 4.40E-15 |
|  | Best | 9.50675E-06 | 0.000168241 | 4.76977E-10 | 0 | 28.96545764 | 7.39532E-05 | 4.70E-17 |
| F34 | Mean | 2.675704311 | 0.285683347 | 0.398662385 | 0.041317662 | 0.000148258 | 0 | 4.04E-01 |
|  | Std | 12.349 | 0.6247 | 1.2164 | 0.1159 | 0.0001 | 0 | 0.2604 |

| | | | | | | | | | |
|---|---|---|---|---|---|---|---|---|---|
| F35 | Best | 0.004253537 | 0.000426605 | 0 | 0.007090687 | 2.28076E-06 | 0 | 1.31E-02 |
| | Mean | 0 | 0 | 0 | 0.002347627 | 0.306523412 | 1.36473E-07 | 0 |
| | Std | 0 | 0 | 0 | 0.0008 | 0.1717 | 0 | 0 |
| | Best | 0 | 0 | 0 | 0.001210743 | 0.018941153 | 9.561E-13 | 0 |
| F36 | Mean | −7684.610475778380 | −12569.48661817300 | −12569.48661817300 | 0.075867128 | 571.8916364 | 0.000949741 | -6.40E+02 |
| | Std | 745.3954 | 0 | 0 | 0.0473 | 334.0422 | 0.0048 | 95.5247 |
| | Best | −8912.885585497820 | −12569.486618173000 | −12569.486618173000 | 0.016965814 | 120.4792858 | 7.63654E-05 | -8.17E+02 |
| F37 | Mean | 0 | 14.56687341 | 0 | 0.01021094 | 0.075818692 | 3.92559E-05 | 1.54E-18 |
| | Std | 0 | 8.7128 | 0 | 0.0082 | 0.0471 | 0.0002 | 1.10E-18 |
| | Best | 0 | 4.042769932 | 0 | 0.004894742 | 0.027562446 | 1E-16 | 1.13E-19 |
| F38 | Mean | 0 | 5E-16 | 0 | 0.067915555 | 0.10357047 | 4.82668E-05 | 2.28E-36 |
| | Std | 0 | 0 | 0 | 0.0789 | 0.0374 | 0.0003 | 3.15E-36 |
| | Best | 0 | 3E-16 | 0 | 0.009036617 | 0.043108935 | 8.45355E-09 | 4.60E-38 |
| F39 | Mean | −10.1061873621653000 | −10.5364098169200000 | −10.5364098166921000 | −10.4028269167718000 | −9.7350533586647200 | −10.5362952172436000 | -1.16E+00 |
| | Std | 1.6679 | 0 | 0 | 0.0004 | 2.0198 | 0.0006 | 6.45E-01 |
| | Best | −10.5364098166921000 | −10.5364098169200000 | −10.5364098166920000 | −10.4029383812040000 | −10.4024580216394000 | −10.5364098166920000 | -4.36E+00 |
| F40 | Mean | −9.5373938082045500 | −10.1531996790582000 | −10.1531996790582000 | −10.1530937574368000 | −8.5402462491687100 | −10.1531978245939000 | -9.80E-01 |
| | Std | 1.9062 | 0 | 0 | 0.0004 | 2.5209 | 0 | 4.88E-01 |
| | Best | −10.1531996790582000 | −10.1531996790582000 | −10.1531996790582000 | −10.1531990814136000 | −10.1522716088073000 | −10.1531996790582000 | -2.70E+00 |
| F41 | Mean | −10.4029405668187000 | −10.4029405668187000 | −10.4029405668187000 | −10.4028269167718000 | −8.5402462491687100 | −10.4028522507988000 | -1.1548 |
| | Std | 0 | 0 | 0 | 0.0004 | 2.5209 | 0.0005 | 0.7481 |
| | Best | −10.4029405668187000 | −10.4029405668187000 | −10.4029405668187000 | −10.4029383812040000 | −10.1522716088073000 | −10.4029405668187000 | -4.4832 |
| F42 | Mean | −186.730907356988000 | −186.730908831024000 | −186.730908831024000 | −186.730844847324000 | −186.728198040376000 | −186.730908827684000 | -137.7200 |
| | Std | 0 | 0 | 0 | 0.0002 | 0.0027 | 0 | 29.1410 |
| | Best | −186.730908831024000 | −186.730908831024000 | −186.730908831024000 | −186.730908739496000 | −186.730889608989000 | −186.730908831024000 | -186.1400 |
| F43 | Mean | −1.0316284534898800 | −1.0316284534898800 | −1.0316284534898800 | −1.0316215131569400 | −1.0254089575785600 | −1.0315698169978900 | -0.0891 |
| | Std | 0 | 0 | 0 | 0 | 0.0098 | 0.0003 | 0.1008 |
| | Best | −1.0316284534898800 | −1.0316284534898800 | −1.0316284534898800 | −1.0316277814769900 | −1.0316217865659200 | −1.0316284534898800 | -1.0300 |
| F44 | Mean | 0 | 4E-16 | 0 | 6.80165E-06 | 0.020952252 | 5.5357E-06 | 1.80E-40 |
|     | Std  | 0      | 0         | 0 | 0           | 0.0288      | 0          | 2.17E-40 |
|     | Best | 0      | 3E-16     | 0 | 3.18002E-09 | 0.000312203 | 0          | 4.36E-42 |
| F45 | Mean | 2.3    | 0         | 0 | 0           | 0           | 0          | 2.51E-02 |
|     | Std  | 1.8597 | 0         | 0 | 0           | 0           | 0          | 1.73E-02 |
|     | Best | 0      | 0         | 0 | 0           | 0           | 0          | 3.50E-03 |
| F47 | Mean | 0      | 5E-16     | 0 | 4.79007E-06 | 0.00452301  | 1.18093E-08 | 0.02656  |
|     | Std  | 0      | 0         | 0 | 0           | 0.0056      | 0          | 0.02526  |
|     | Best | 0      | 3E-16     | 0 | 4.85008E-09 | 0.000116302 | 0          | 3.70E-05 |
| F50 | Mean | 0      | 4.0238E-08 | 0 | 0.00084717  | 0.099715402 | 2.02917E-05 | 4.05E-39 |
|     | Std  | 0      | 0         | 0 | 0.0003      | 0.0472      | 0.0001     | 6.50E-39 |
|     | Best | 0      | 2.1E-14   | 0 | 0.000614477 | 0.026318335 | 5.25441E-09 | 1.05E-40 |

The following functions (F1, F3, F4, F12, F15, F16, F17, F22, F28, F29, F31, F32, F33, F46, F48, F49) were also tested using ACO-CI however, results were not satisfactory. The results obtained through where comparative study of algorithms such as PSO, ABC, BSA, GA, CI, ARGA and ACO-CI was done is mentioned in Table 1. The outcome consists of mean, standard and best solution for 34 benchmark functions and each function has been assessed for generating 30 outputs in total to obtain a normalized figure.

**4. Test examples**

In the present work, the ACO-CI hybrid algorithm was successfully applied for solving two continuous variable mechanical design engineering optimization problems. These problems are well studied in the literature and used to compare the performance of various optimization algorithms such as Cuckoo search (CS), Symbiotic organisms search (SOS), Colliding bodies optimization (CBO), cohort intelligence with Self adaptive penalty function (CI-SAPF), Cohort intelligence with Self adaptive penalty function with Colliding bodies optimization (CI-SAPF-CBO). Furthermore, for every individual problem ACO-CI was solved 30 times with different initialization. The mathematical formulation, results and comparison of solution with other contemporary algorithms are discussed in the following sections. The specially developed ACO-CI hybrid algorithm has been successfully applied to solve the mechanical engineering design problems.

**Stepped Cantilever Beam**

The square cross-section stepped cantilever beam's (refer Figure 2) weight optimization is the subject of the issue. At one end, the beam is fixed, while force is applied at the other. The thickness is maintained constant in this issue (here, t = 2/3), and the variables are the heights (or widths) of the various beam components. $0.01 \leq x_i \leq 100$ are the set bound limitations. Analytically, this issue may be stated as follows (Gandomi et al., 2013):

$$\text{Minimize: } f(x) = 0.0624(x_1 + x_2 + x_3 + x_4 + x_5) \quad (10)$$

$$\text{Subject to: } g(x) = \frac{61}{x_1^3} + \frac{37}{x_2^3} + \frac{19}{x_3^3} + \frac{7}{x_4^3} + \frac{1}{x_5^3} - 1 \leq 0$$

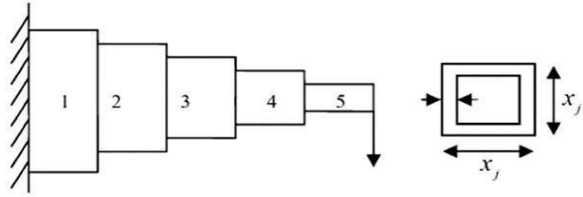

**Fig. 2** Stepped cantilever beam (Gandomi, Yang and Alavi, 2013)

The total weight reduction is the goal of the cantilever beam design problem consisting of continuous variables. This problem's resolution and comparison to other modern algorithms in the Table 2 satisfactorily validate the ACO-CI algorithm. The following lists the solutions provided by ACO-CI. The function values used by ACO-CI to solve the cantilever beam issue are extremely comparable to those used by CS and CI-SAPF-CBO and are just as reliable.

**Table. 2** Comparative results of ACO-CI with CS, SOS, CBO, CI-SAPF, CI-SAPF-CBO for stepped cantilever beam

| Techniques | CS (Gandomi, et al., 2013) | SOS (Cheng and Prayogo, 2014) | CBO (Kale and Kulkarni, 2021) | CI-SAPF (Kale and Kulkarni, 2021) | CI-SAPF-CBO (Kale and Kulkarni, 2021) | ACO-CI |
|---|---|---|---|---|---|---|
| **Min. weight** | 1.3400 | 1.3400 | 3.2000 | 1.3400 | 1.3400 | 1.3399 |
| **Function evaluations** | NA | 15000 | 2190 | 13750 | 3025 | 19339.0900 |
| $X_1$ | 6.0089 | 6.0188 | NA | NA | NA | 6.0082 |
| $X_2$ | 5.3049 | 5.3034 | NA | NA | NA | 5.3229 |
| $X_3$ | 4.5023 | 4.4959 | NA | NA | NA | 4.4879 |
| $X_4$ | 4.5023 | 3.4990 | NA | NA | NA | 3.5039 |
| $X_5$ | 2.1504 | 2.1556 | NA | NA | NA | 2.1509 |

The ACO-CI hybrid algorithm offered a very competitive end result when it was applied on the Stepped Cantilever Beam problem in comparison to the other optimization algorithm. This is clearly depicted in the Table 1 where the results from ACO-CI and other algorithms are compared for the described problem. The resulting value obtained from the ACO-CI hybrid algorithm was 1.339941 which is comparatively less than the other optimization algorithm. This settles that our ACO-CI hybrid algorithm performed better than the CS, SOS, CBO, CI-SAPF and CI-SAPF-CBO optimization algorithm in terms of minimizing the weight. The values of the variables are also compared in the Table 1.

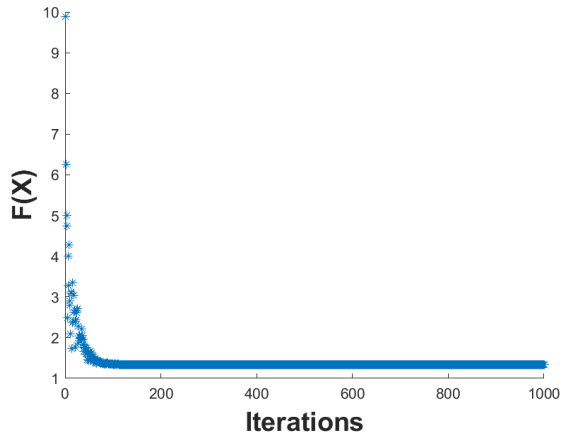

**Fig 3.** Convergence graph of stepped cantilever beam

The convergence graph has been plotted between the Function value and the number of iterations for the stepped cantilever beam problem (refer Figure 3).

### 3.1 I-Section Beam

The objective function is minimization of vertical deflection of I-section beam which can be formulated as follows, (Ayrupa *et al.*, 2019):

$$F(x) = \frac{PL^3}{48EI} \tag{11}$$

$$I = \frac{t_w(h - 2t_f)^3}{12} + \frac{bt_f^3}{6} + 2bt_f\left(\frac{h - t_f}{2}\right)^2 \tag{12}$$

The ranges of beam dimensions, which are design parameters belonging to problem, are as follows:

$$10 \leq h \leq 100 \tag{13}$$

$$10 \leq b \leq 60 \tag{14}$$

$$0.9 \leq t_w \leq 6 \tag{15}$$

$$0.9 \leq t_f \leq 6 \tag{16}$$

The design constraints are $g_1$ and $g_2$; respectively. They express that beam section may not be bigger than 300 cm² and allowable moment stress may not be bigger than 6 $kN/cm^2$ with equations shown as:

$$g_1 = 2bt_f + t_w(h - 2t_f) \leq 300 \tag{17}$$

$$g_2 = \frac{1.5PLH}{t_w(h - 2t_f)^3 + 2bt_w(4t_f^2 + 3h(h - 2t_f))} + \frac{1.5QLb}{t_w^3(h - 2t_f) + 2t_w b^3} \leq 6 \tag{18}$$

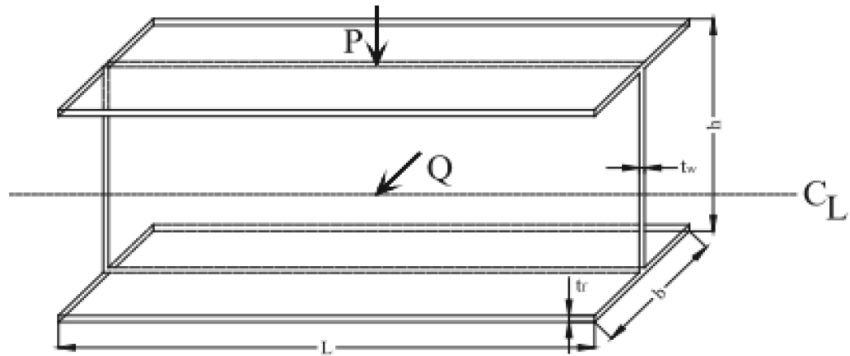

**Fig. 4** I-section beam (Ayrupa et al., 2019)

**Table 3** Comparative results of ACO-CI with ANN for I-section beam problem

| CASES | Length(L) | Load(P) | ANN min values (Ayrupa et al., 2019) | ACO CI min values |
|---|---|---|---|---|
| CASE 1 | 120 | 652 | 0.002018 | 0.002018 |
| CASE 2 | 350 | 520 | 0.049381 | 0.049381 |
| CASE 3 | 285 | 743 | 0.038774 | 0.038774 |
| CASE 4 | 150 | 200 | 0.001209 | 0.001209 |
| CASE 5 | 345 | 264 | 0.020572 | 0.020572 |
| CASE 6 | 100 | 690 | 0.001236 | 0.001235 |
| CASE 7 | 250 | 442 | 0.012915 | 0.012915 |
| CASE 8 | 310 | 675 | 0.049937 | 0.045771 |
| CASE 9 | 270 | 482 | 0.018465 | 0.018464 |
| CASE 10 | 220 | 355 | 0.006771 | 0.006771 |

The ACO-CI algorithm was applied for the I-section beam problem. This ACO-CI algorithm was compared with the ANN model. Different values of load and length were tested, and ACO-CI performed with much better results compared to the ANN model. The best values obtained from ACO-CI for each case are presented in the Table 2 For the I section beam problem, the results that were obtained by applying the ACO-CI algorithm have been compared with the results obtained by other method and are shown in the above Table 2. 10 cases have been considered from which the horizontal length (L) of I section beam and vertical load (P) on the beam are variable. In each case it can be seen that most of the values obtained through ACO-CI are nearly equal and few are even less. This basically shows us that the results obtained from the ACO-CI hybrid algorithm are better compared to the ANN method results.

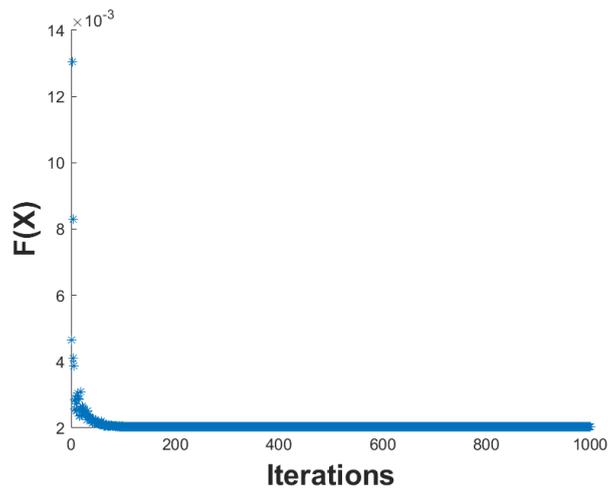

**Fig 5.** Convergence graph for the I section beam problem

The convergence graph is plotted above between the function value and the number of iterations for the given I section beam problem.

**Table 4** Statistical results obtained by ACO-CI for stepped cantilever beam problem and I section beam problem

| PROBLEM | CASE | MEAN | BEST | STANDARD DEVIATION | WORST | AVERAGE TIME COMPUTE | AVERAGE NO OF ITERATION | FUNCTION EVALUATION |
|---|---|---|---|---|---|---|---|---|
| 1 | | 1.3400 | 1.3399 | 7.42E-05 | 1.3402 | 2.8460 | 641.9333 | 19339.0900 |
| 2 | 1 | 0.0020 | 0.0020 | 7.22E-06 | 0.0020 | 2.5177 | 594.7333 | 17108.1800 |
| | 2 | 0.0493 | 0.0493 | 3.62E-05 | 0.0495 | 3.1036 | 703.0667 | 21090 |
| | 3 | 0.0389 | 0.0387 | 0.0005 | 0.0412 | 3.8767 | 879.2667 | 26342.7300 |
| | 4 | 0.0012 | 0.0012 | 6.54E-06 | 0.0012 | 3.5351 | 797.6000 | 24021.8200 |
| | 5 | 0.0205 | 0.0205 | 1.52E-05 | 0.0206 | 3.5479 | 805.1333 | 24109.0900 |
| | 6 | 0.0012 | 0.0012 | 3.54E-07 | 0.0012 | 2.9339 | 668.1333 | 19936.3636 |
| | 7 | 0.0129 | 0.0129 | 1.53E-07 | 0.0129 | 3.1819 | 721.4000 | 21621.8181 |
| | 8 | 0.0462 | 0.0457 | 9.06E-04 | 0.0492 | 3.4011 | 770.3333 | 23110.9090 |
| | 9 | 0.0184 | 0.0184 | 8.37E-08 | 0.0184 | 3.2477 | 735.5333 | 22069.0909 |
| | 10 | 0.0067 | 0.0067 | 1.27E-06 | 0.0067 | 3.4396 | 776.6666 | 23372.7272 |

Proper validation of the result that was obtained from the ACO-CI hybrid algorithm was done from which the output is shown above in the Table 3. The ACO-CI algorithm was applied on the cantilever beam problem and 30 such outputs were generated. From those 30 outputs, it was observed that the best value was at 1.339941 with the mean value being 1.340047, worst being 1.340236 and the standard deviation of 0.0000742. The convergence of value was received on an average at the 640th iteration with the computational time of 2.846024 seconds.

The results that were obtained when ACO-CI hybrid algorithm was applied on the I section beam problem are discussed in the Table 3. The results basically comprise of 10 cases, each having different values for load and the horizontal length of the I section beam. Each case has been tested out and 30 outputs were generated. From these 30 outputs, the best value, standard deviation, worst value, average computational time, average number of iteration and function evaluation were obtained and this process was done for all the 10 cases.

5. **Conclusion**

The ability of ACO-CI is depicted over here to solve the continuous variable constrain problem. The penalty function approach is adapted to for constrain handling. The following paper made use of ACO-CI algorithm to solve mechanical design problems. The Algorithm is validated by solving I section beam design problem and stepped cantilever beam problem. The I section beam problem consisted of 4 variables and 2 constraints and the stepped cantilever beam problem consisted of 5 variables and 1 constraint. From the results analysis and comparison, it is noticed that ACO CI algorithm performed better in obtaining robust solutions. The ACO-CI algorithm is hybridized by adopting the prominent qualities of ACO and CI algorithm. Finally, the algorithm is tested on benchmark problems to check statistical significance of ACO-CI for all 50 problems considered. The successfully created ACO-CI hybrid now can be used to solve various real world mechanical design problems.

**Conflict of Interest:**

We wish to confirm that there are no known conflicts of interest associated with this publication and there has been no significant financial support for this work that could have influenced its outcome. We confirm that the manuscript has been read and approved by all named authors and that there are no other persons who satisfied the criteria for authorship but are not listed. We further confirm that the order of authors listed in the manuscript has been approved by all of us.